# Développement automatique de lexiques pour les concepts émergents : une exploration méthodologique

Revekka Kyriakoglou[1], Anna Pappa[1], Jilin He[1], Antoine Schoen[2], Patricia Laurens[2], Markarit Vartampetian[3], Philippe Laredo[2], Tita Kyriacopoulou[3]

[1]LIASD - Université Paris 8, – kyriakoglou@up8.edu, ap@up8.edu, patrick.he1024@gmail.com

[2]LISIS - Université Gustave Eiffel, – antoine.schoen@esiee.fr, patricia.laurens@esiee.fr, philippe.laredo@univ-eiffel.fr,

[3]LIGM - Université Gustave Eiffel, – tita@univ-mlv.fr, markaritvar@gmail.com,

## Abstract

This paper presents the development of a lexicon centered on emerging concepts, focusing on *non-technological innovation*[1]. It introduces a four-step methodology that combines human expertise, statistical analysis, and machine learning techniques to establish a model that can be generalized across multiple domains. This process includes the creation of a thematic corpus, the development of a *Gold Standard Lexicon*, annotation and preparation of a training corpus, and finally, the implementation of learning models to identify new terms. The results demonstrate the robustness and relevance of our approach, highlighting its adaptability to various contexts and its contribution to lexical research. The developed methodology promises applicability in conceptual fields.

## Résumé

Cet article présente le développement d'un lexique centré sur les concepts émergents, en se focalisant sur *l'innovation non technologique*. Il décrit une méthodologie en quatre étapes, combinant expertise humaine, statistiques et techniques d'apprentissage automatique, pour établir un modèle généralisable à plusieurs domaines. Cette procédure comprend la création d'un corpus thématique, la constitution d'un lexique de référence, l'annotation et la préparation d'un corpus d'entraînement, et enfin, l'implémentation de modèles d'apprentissage pour identifier de nouveaux termes. Les résultats montrent la robustesse et la pertinence de notre approche, mettant en évidence sa capacité à être adaptée à plusieurs contextes et sa contribution à la recherche lexicale. La méthodologie développée promet une applicabilité dans des domaines conceptuels.

**Mots clés :** développement de lexique sémantique, modélisation de domaine conceptuel, identification de concepts émergents, extraction de connaissances, annotation par apprentissage automatique, création de corpus.

## 1. Introduction

Dans le domaine en évolution rapide de l'analyse des données textuelles (ADT[2]), notre étude se concentre sur une approche systématique visant à décrire le contenu textuel pour identifier et représenter des concepts, qu'ils soient émergents ou déjà établis. La problématique de notre recherche s'articule autour de la capacité à détecter des termes nouveaux qui désignent des concepts spécifiques à un domaine donné, un défi majeur face aux méthodes traditionnelles d'extraction terminologique (L'Homme, 2004 ; Pazienza et al., 2005). Ces dernières, s'appuyant principalement sur l'analyse statistique, se concentrent sur la fréquence et la co-occurrence des termes, ce qui peut limiter leur capacité à garantir la nouveauté des termes ou leur adéquation précise avec le domaine concerné[3]. Cette limite soulève un point critique, d'autant plus que, du point de vue de la recherche qualitative, les contributions méthodologiques restent insuffisantes malgré l'existence de résultats prometteurs. Face à ce contexte, notre recherche présente une méthodologie en quatre étapes, conçue pour développer des lexiques centrés sur l'identification de concepts émergents, avec cas d'étude l'innovation non technologique. Cette démarche s'inscrit dans une perspective de l'ADT qualitative, visant à relever le défi de l'extraction de termes nouveaux ou spécifiquement liés à un domaine, un objectif déjà souligné par des travaux antérieurs (Gillam et al., 2005; Avancini et al., 2006; Kazi et al., 2014; Pasini et Camacho-Collados, 2020). Pour cela, nous avons constitué un corpus spécifique aux contenus associés à cinq mots-clés recommandés par des experts, couvrant plusieurs secteurs économiques. À l'aide de la plateforme *Cortext*[4], un lexique de référence comprenant 306 termes a été établi, permettant d'entraîner et évaluer nos modèles d'apprentissage automatique. Une sélection de ce corpus a été annotée pour identifier les contextes des termes référencés, formant des blocs de contexte pour l'entraînement de nos modèles, une démarche qui met en évidence l'efficacité des modèles d'apprentissage profond dans l'automatisation de l'annotation et leur contribution à la recherche linguistique et sociale (Ruoyu et al., 2023). Notre méthode dépasse les limites des analyses statistiques traditionnelles, comme les approches Rake (Rose et al., 2010) ou Yake (Campos et al., 2020), en intégrant une expertise humaine essentielle pour la validation conceptuelle des termes. Ce processus non seulement aborde les défis liés au manque de corpus annotés et aux difficultés d'application des algorithmes d'apprentissage pour l'extraction de concepts émergents (Hasan et Ng, 2014) mais souligne également l'importance de l'expertise humaine dans la validation des termes. Le plan de cet article se déploie comme suit : après cette introduction, nous développerons chacune des quatre phases de notre méthodologie dans les sections suivantes, avant de conclure sur l'efficacité de notre approche et ses contributions à l'ADT qualitative et à l'identification de concepts émergents dans le domaine de l'innovation non technologique. Cette structure vise à fournir un cadre pour comprendre notre démarche et les résultats obtenus[5], établissant un lien direct entre la méthodologie proposée et son application dans le domaine de l'analyse des données textuelles.

---

[1] L'*innovation non technologique* englobe les changements dans les domaines de l'organisation, du management, des pratiques de travail, de la stratégie d'entreprise, du marketing, ou encore des modèles d'affaires apportant une valeur ajoutée sans nécessairement s'appuyer sur de nouvelles technologies (Tidd et Bessant, 2020).

[2] Cette notion désigne l'ensemble des "approches, méthodes et outils informatiques visant à découvrir l'information contenue dans un corps textuel" (Boughzala et al., 2014).

## 2. Création de corpus

Pour développer des ressources dynamiques qui reflètent fidèlement les connaissances du monde réel, la création d'un grand corpus multilingue est essentielle. Dès lors, le *World Wide Web*, considéré comme une riche source de données depuis Kilgarriff et Grefenstette (2003), sert de fondement à notre corpus. Notre corpus[6] est principalement constitué de textes décrivant les initiatives d'innovation d'entreprises, récupérés à la fois de leurs sites internet et de leurs rapports annuels, grâce à une liste de 4000 entreprises réparties sur 27 secteurs économiques, fournie par des spécialistes. Cette démarche a permis de distinguer deux ensembles de données : un corpus web, issus des sites d'entreprises, et un corpus PDF, constitué de rapports annuels. Cette approche enrichit notre corpus à travers différentes sources documentaires. Nous utilisons *Scrapy* (Myers et McGuffee, 2015) pour cibler des sections pertinentes via un analyseur de balises dans l'arborescence DOM[7], éliminant ainsi le bruit et le contenu redondant. Notre méthode initie la création de corpus avec un ensemble de cinq 'mots-clés conceptuels' déterminés par des experts, visant à capturer des textes liés à l'innovation depuis les sites d'entreprises, en se concentrant sur ces cinq termes.

---

[3] Plusieurs outils proposés peuvent être testés sur https://www.istex.fr/etat-de-lart-des-outils-dextraction-terminologique/

[4] https://www.cortext.net/

[5] Nous mettons à disposition le code et les modèles utilisés : https://code.up8.edu/glacon/malantin.git.

[6] Notre corpus est composé de textes écrits en plusieurs langues mais pour notre cas d'étude, nous n'utilisons que le corpus en anglais.

### 2.1. Corpus web

La phase de création de notre corpus web a impliqué l'application des filtres sélectifs sur les URL et le contenu textuel pour isoler des informations pertinentes. Les URL ont été filtrées pour cibler la page d'accueil des entreprises, excluant celles contenant des termes non pertinents afin de réduire le bruit. Seuls les paragraphes HTML (<p>) et les balises <title>, <h1>, et <h2> contenant les 5 termes clés ont été conservés favorisant ainsi la qualité et la pertinence des données collectées. Concernant la diversité linguistique, nous avons adopté une approche en deux étapes pour vérifier la langue utilisant l'attribut <lang> puis, pour une vérification plus approfondie, la bibliothèque *googletrans*. Cette stratégie nous a permis de composer un corpus multilingue riche et varié, provenant de 924 sites web analysés, comme on voit dans le Tableau 1, démontrant ainsi la diversité linguistique et sectorielle des ressources web pertinentes.

### 2.2. Corpus pdf

Pour élaborer le corpus PDF, nous avons développé un parseur utilisant *libcurl*, pour télécharger les contenus des rapports annuels, avec un filtrage précis des URL selon des critères prédéfinis. Ce processus s'appuie sur un dictionnaire des noms d'entreprises pour effectuer des recherches ciblées et récupérer les documents pertinents. En excluant les contenus non désirés comme .doc et .RTF, et en se concentrant sur les données officielles et structurées, nous avons recueilli 8368 rapports annuels de 2017 à 2021, à partir de *annualreports.com*, assurant une représentation diversifiée des secteurs économiques.

## 3. Création du "lexique de référence"

Pour développer le "*lexique de référence*", nous avons utilisé la plateforme *Cortext* et un corpus web en appliquant les principes d'unithéité et de terméité[8] (Frantzi et al., 2000) pour filtrer environ 20 000 articles (Figure 1). Cela a conduit à la création d'un lexique composé de 306 termes répartis en six catégories cognitives liés à l'innovation : *durabilité (sus), transformation numérique (dig), gestion du changement (mag), activités d'innovation (inn), modèles d'entreprise (bus) et la responsabilité sociale des entreprises (cor)*. Cette méthode basée sur l'analyse de sites de 3992 entreprises, a abouti à la définition d'un *lexique de référence* suivant une stratégie rigoureuse pour capturer et structurer les connaissances dans le domaine de l'innovation, spécifiquement orientée vers l'innovation non technologique.

---

[7] DOM (Document Object Model) https://www.w3.org/TR/WD-DOM/introduction.html

[8] Les notions d'unithéité (*unithood*) et terméité (*termhood*) font référence respectivement à la cohésion d'une expression multi-mots pour former un concept unique et au degré auquel une expression est associée à un domaine de connaissances spécifiques (Kageura et Umino, 1996).

REVEKKA KYRIAKOGLOU, ANNA PAPPA, JILIN HE, ANTOINE SCHOEN, PATRICIA LAURENS, MARKARIT
VARTAMPETIAN, PHILIPPE LAREDO, TITA KYRIACOPOULOU

| Total 3835 | Secteur d'activités des entreprises |
|---|---|
| 1086 | Machinerie électrique, électronique industrielle |
| 634 | Chimie, pétrole, caoutchouc et plastique |
| 387 | Services aux entreprises |
| 202 | Fabrication de matériel de transport |
| 183 | Communications |
| 158 | Métaux et produits métalliques |
| 141 | Logiciels informatiques |
| 111 | Commerce de gros |
| 108 | Fabrication de produits alimentaires, tabac |
| 101 | Banque, assurances et services financiers |
| 91 | Biotechnologie et sciences de la vie |
| 81 | Matériel informatique |
| 63 | Exploitation minière et extraction |
| 57 | Services d'utilité publique |
| 49 | Produits en cuir, pierre, argile et verre |
| 48 | Fabrication de meubles |
| 47 | Construction |
| 44 | Médias et diffusion |
| 43 | Fabrication de textiles et de vêtements |
| 40 | Commerce de détail |
| 36 | Transport, fret et stockage |
| 33 | Fabrication diverse |
| 31 | Voyages, personnel et loisirs |
| 22 | Administration publique |
| 22 | Imprimerie et édition |
| 10 | Agriculture, horticulture et élevage |
| 7 | Services immobiliers |

| Langues | URLs | Secteurs | Tokens |
|---|---|---|---|
| Anglais | 41224 | 27 | 8133370 |
| Chinois | 33054 | 22 | 1823235 |
| Français | 15074 | 17 | 2821719 |
| Allemand | 16783 | 16 | 3327026 |
| Espagnol | 10364 | 12 | 2483900 |
| Italien | 8236 | 12 | 1323276 |
| Coréen | 10929 | 11 | 1899763 |
| Néerlandais | 1907 | 11 | 505163 |
| Catalan | 2427 | 9 | 11491180 |
| Portugais | 7594 | 8 | 780462 |
| Danois | 1088 | 8 | 234798 |
| Russe | 32088 | 7 | 13157829 |
| Suédois | 1355 | 7 | 219159 |
| Polonais | 1632 | 6 | 241973 |
| Finlandais | 1029 | 4 | 99361 |
| Ukrainien | 3073 | 1 | 981 |

*Tableau 1 : (à gauche) Répartition des entreprises par secteur d'activité. (à droite) Répartition des URLs, secteurs, et tokens par langue, filtrée par les mots-clés : innovation, recherche, development, strategy, design*

## 4. Annotation et préparation du corpus

L'annotation du corpus est réalisée en plusieurs étapes clés. Le découpage et étiquetage des phrases sont effectués à l'aide de la librairie *nltk*. La recherche des phrases contenant des termes du lexique de référence se fait au sein du corpus. L'identification d'un terme conduit systématiquement à une vérification préalable, s'assurant qu'il ne s'agit pas d'un terme englobant qui pourrait entraîner des correspondances partielles, comme le fait d'identifier '*innovation*' dans '*innovation technology*'. L'étiquetage du terme détecté se réalise en lui attribuant l'étiquette <mot>, et l'intégration de sa catégorie spécifique comme attribut (de cette étiquette) renforce la précision de l'annotation, comme dans l'exemple qui suit :

*<phrase category='Digital transformation' values='virtual reality'> As part of our portfolio of precision healthcare solutions, we offer 3D <mot category='Digital transformation'>virtual reality</mot> simulators and simulator modules for medical applications.</phrase>*

Cet exemple illustre la manière dont les termes du lexique sont identifiés et étiquetés au sein d'une phrase, en soulignant l'importance de l'attribution de catégories pour améliorer la qualité et la pertinence de l'annotation du corpus. Ces catégories dénotent divers aspects ou types d'innovation. Pour assurer l'uniformité des données, les termes sont normalisés à leur forme canonique.

Afin de créer un jeu de données contextuellement riche, nous intégrons les termes étiquetés directement dans leurs phrases, d'origine. Cette approche garantit que le jeu de données fournit aux modèles d'apprentissage automatique les informations nécessaires pour non seulement détecter la présence de termes mais aussi saisir leur signification sémantique dans les contextes d'innovation non-technologique.

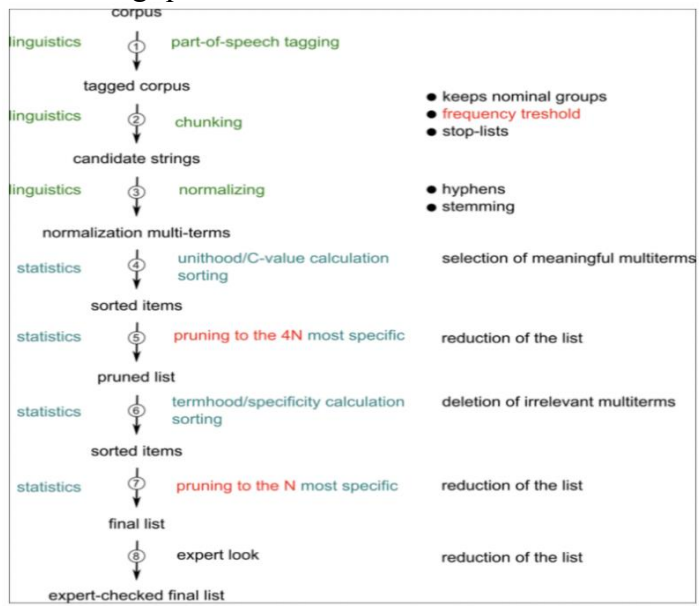

*Figure 1 : Création du lexique de référence*

Pour délimiter le contexte autour des termes pertinents, nous créons des blocs textuels de longueur variable à partir des phrases, centrés sur une phrase ayant un terme du lexique et incluant les deux phrases adjacentes de chaque côté>, comme illustré ci-dessous :

*<phrase>Sentence 1</phrase><phrase> Sentence 2</phrase><phrase category='I-XXX' values=term> Sentence with term<mot category='I-XXX'> term </mot> Sentence with term</phrase><phrase> Sentence 4</phrase><phrase> Sentence 5</phrase>*

Pour approfondir l'analyse sémantique, la similarité cosinus est employée afin d'évaluer la proximité thématique des mots-clés avec le bloc concerné, fixant un seuil de 0.5 pour cette mesure. Seuls les blocs contenant les mots-clés sémantiquement alignés sont conservés, renforçant ainsi la précision et la pertinence de notre corpus annoté.

L'approche d'étiquetage adoptée, tient compte de la structure *multi-tokens* de la majorité des termes de notre 'lexique de référence', qui varie entre de deux à quatre *tokens*. Pour cela, nous employons le forma BIO[9], qui permet une distinction nette du début (B), de l'intérieur (I) et de l'extérieur (O) des entités dans le texte. Cette méthode de balisage introduite par Ramshaw et Marcus (1999) est devenue un standard dans le domaine du traitement du langage naturel (NLP), comme en témoignent les travaux de Sang et De Meulder (2003) ainsi que de Gu et al., (2021), en raison de son efficacité à gérer les séquences de *tokens* et à faciliter l'identification précise des entités.

Considérons une phrase $s = \{w_1, w_2, ..., w_i, ..., w_n\}$ composée de $n$ mots, et un mot-clé *multi-tokens* apparaissant à la position $i$ dans la phrase. Nous utilisons la notation $kw = \{w_i, ..., w_{i+n}\}$ où n est le nombre de *tokens* du mot-clé. La tâche consiste à effectuer un étiquetage séquentiel sur la phrase pour extraire le mot-clé cible et la catégorie à laquelle il appartient. Les mots étiquetés avec 'O' ne sont pas des mots-clés et l'étiquette 'I-XXX' est utilisée pour les mots qui sont des mots-clés de la catégorie 'XXX'. Lorsque deux entités de type 'XXX' sont immédiatement adjacentes, le premier mot de la seconde entité est marqué 'B-XXX', indiquant qu'il initie une nouvelle entité. Dans le cas de recouvrement de mots-clés, par exemple le terme *'fonctionnalité et solutions de design'* qui est le chevauchement des termes *'fonctionnalité et design'* et '*solutions de design*' présents dans le lexique de référence, nous introduisons une nouvelle catégorie appelée 'I-mac' pour décrire ce phénomène que nous appelons '*macro-terme*'.

---

[9] Le format BIO (*Begin, Inside, Outside*) est une méthode de balisage, utilisée pour l'annotation des séquences.

| Corpus | Entraînement | Test |
|--------|--------------|------|
| 1 | 100% corpus web (100% mots-clés)<br>16241 blocs | 100% corpus pdf<br>27735 blocs |
| 2 | ∼80% corpus web (80% mots-clés)<br>11107 blocs | 100% corpus pdf + ∼20% corpus web<br>32869 blocs |
| 3 | ∼75% corpus web + ∼25% corpus pdf (∼70% mots-clés)<br>16637 blocs | ∼25% corpus web + ∼75% corpus pdf<br>27405 blocs |
| 4 | ∼50% corpus web + ∼50% corpus pdf (50% mots-clés dans le web et 50% dans le pdf)<br>16745 blocs | ∼50% corpus web et ∼50% corpus pdf<br>27321 blocs |

*Tableau 2 : Création des jeux de données*

Pour affiner les performances des modèles d'apprentissage automatique, nous avons élaboré quatre jeux de données distincts à partir de nos corpus web et PDF comme détaillé dans le Tableau 2. Ces ensembles visent à capturer une gamme étendue de contextes liés aux *innovations non technologiques*, permettant aux modèles de mieux saisir la diversité et la complexité du langage utilisé dans le domaine. Ces jeux de données servent ensuite à entraîner des modèles pour identifier et classifier les termes relatifs à l'innovation non technologique.

## 5. Apprentissage de modèles et identification de nouveaux termes

Dans notre recherche, nous avons exploré quatre modèles distincts pour le traitement et l'analyse de données textuelles, visant à maximiser la précision dans l'identification de nouveaux termes liés à l'innovation non technologique. Nous avons déployé un réseau neuronal convolutif (CNN) et l'avons étendu à un modèle CNN-CRF en intégrant une couche de champ aléatoire conditionnel (CRF) ainsi que deux architectures basées sur les transformateurs : BERT (Devlin et al., 2018) et RoBERTa (Liu et al., 2019).

Le CNN débute avec une couche de plongement transformant les *tokens* d'entrée en vecteurs de 300 dimensions, adaptés aux séquences de longueurs variables grâce à un mécanisme du *padding* et se concentre uniquement sur les *tokens* présents au moins trois fois (min_freq = 3) dans le corpus. Cette architecture progresse à travers des couches convolutionnelles parallèles, avec des noyaux de taille trois et cinq, produisant chacune 128 caractéristiques. Les caractéristiques extraites sont ensuite fusionnées et affinées, à travers trois couches convolutionnelles supplémentaires, chacune avec 256 caractéristiques en sortie et une taille de noyau de cinq. Ces couches utilisent la fonction d'activation Unité Linéaire Rectifiée (ReLU) et sont entrecoupées de *dropout* à un taux de 0.5 pour atténuer le surajustement.

Le modèle CNN mènent à une couche entièrement connectée qui transforme les caractéristiques de haut niveau dérivées du traitement convolutionnel pour correspondre à l'espace de sortie cible. Cette transformation est médiée soit par une couche de champ aléatoire conditionnel (CRF), formant ainsi le modèle CNN-CRF, soit par une fonction log *softmax*, en fonction de la configuration du modèle. Pour les modèles équipés de la couche CRF, la couche de sortie génère des scores en utilisant l'algorithme de Viterbi du CRF pour le décodage de la séquence de tags la plus probable lors de l'inférence ou calcule la log-vraisemblance négative de la séquence de tags correcte pendant l'entraînement. En l'absence d'une couche CRF, le modèle applique une fonction log *softmax* à travers la dernière dimension des *logits* pour produire une distribution de probabilité sur les classes cibles pour chaque *token*. Pendant l'entraînement, cette distribution de probabilité est utilisée pour calculer la perte de log-vraisemblance négative, en tenant compte des vraies étiquettes de classe et des longueurs de séquence pour les données d'entrée en lot.

Les architectures BERT et RoBERTa ont été spécifiquement ajustées pour la classification de *tokens* en fonction des catégories définies par notre 'lexique de référence'. Ces modèles prédictifs ont été entraînés pour reconnaître et attribuer correctement les catégories aux *tokens*, enrichissant notre compréhension des termes liés à l'innovation non technologique.

Suite à l'entraînement, l'identification des termes significatifs repose sur la mesure de similarité cosinus adoptant le même seuil de 0.5 utilisé pour la constitution des jeux de données. Cette étape vise à distinguer les expressions pertinentes, excluant les *mono-termes*, c'est-à-dire les termes constitués d'un seul mot, pour leur manque de spécificité (cycle, effort, équipement, dépenses, vie, modèle, produit, ...), et ainsi affiner la sélection des termes révélateur du contexte d'innovation technologique.

## 6. Résultats

L'analyse comparative des performances des modèles CRF, RoBERTa, et les autres architectures, relève des observations clés concernant leur efficacité dans le traitement de données textuelles. Grâce à des modèles basés sur BERT, nous avons obtenu des scores F1 de 73% et des taux de précision de 76% dans l'identification des termes du lexique de référence.

Selon le Tableau 3, le modèle CRF se distingue par sa précision élevée sur tous les jeux de données, comme le montrent les scores en gras, et affiche une performance globalement équilibrée selon le score F1, qui mesure l'harmonie entre la précision et le rappel. Malgré cela RoBERTa obtient le score le plus élevé et surpasse significativement les autres modèles dans l'identification de nouveaux termes liés à des concepts, avec un taux d'acceptation moyen de 67% (Tableau 4). RoBerta obtient le rappel le plus élevé aussi (Figure 2). Cette performance avancée peut être attribuée à l'architecture d'apprentissage en profondeur de RoBERTa, qui est capable de comprendre le contexte et la sémantique d'un texte. Une telle compréhension permet à RoBERTa d'identifier de nouveaux termes que les modèles traditionnels pourraient négliger, ce qui souligne sa capacité à s'adapter aux concepts émergents spécifiques à un domaine.

**Précision**

| Modèle | Jeu de données 1 | Jeu de données 2 | Jeu de données 3 | Jeu de données 4 | Moyenne |
|--------|-----------------|-----------------|-----------------|-----------------|---------|
| CNN | 0.8056 | 0.7445 | 0.8174 | 0.8589 | 0.8066 |
| CRF | **0.8286** | **0.8701** | **0.8938** | **0.8796** | **0.8680** |
| BERT | 0.7769 | 0.7435 | 0.8632 | 0.7957 | 0.7948 |
| RoBERTa | 0.7801 | 0.7943 | 0.8405 | 0.6576 | 0.7681 |

**Rappel**

| Modèle | Jeu de données 1 | Jeu de données 2 | Jeu de données 3 | Jeu de données 4 | Moyenne |
|--------|------------------|------------------|------------------|------------------|---------|
| CNN | 0.7417 | 0.7126 | 0.5794 | **0.6523** | 0.6715 |
| CRF | 0.7406 | 0.7179 | 0.5930 | 0.6479 | 0.6749 |
| BERT | 0.7667 | 0.7326 | 0.5885 | 0.6465 | 0.6836 |
| RoBERTa | **0.7826** | **0.7391** | **0.6734** | 0.6052 | **0.7001** |

**F1-Score**

| Modèle | Jeu de données 1 | Jeu de données 2 | Jeu de données 3 | Jeu de données 4 | Moyenne |
|--------|------------------|------------------|------------------|------------------|---------|
| CNN | 0.7723 | 0.7282 | 0.6781 | 0.7415 | 0.7300 |
| CRF | **0.7821** | **0.7867** | 0.7130 | **0.7462** | **0.7570** |
| BERT | 0.7718 | 0.7380 | 0.6999 | 0.7133 | 0.7308 |
| RoBERTa | 0.7814 | 0.7657 | **0.7477** | 0.6304 | 0.7313 |

*Tableau 3 : Performance des modèles à travers les jeux de données*

À l'inverse, le modèle CRF excelle dans sa précision lorsqu'il traite des termes connus provenant du lexique de référence, démontrant ainsi sa capacité à annoter avec précision des textes avec un vocabulaire bien défini. Cependant, son efficacité diminue lorsqu'il s'agit d'identifier de nouveaux termes, puisque seulement 25 % de ces nouveaux termes sont validés par les experts. Cela peut s'expliquer par la dépendance de la CRF à des caractéristiques définies manuellement et par son approche structurée de l'apprentissage et de la prédiction. Contrairement aux modèles d'apprentissage profond, le CRF semble incapable de saisir la dynamique contextuelle essentielle à la reconnaissance de nouveaux termes.

En ce qui concerne les jeux de données, le jeu de données 2 semble avoir des scores de rappel plus faibles, ce qui pourrait suggérer qu'il est plus difficile pour les modèles de capturer toutes les instances pertinentes. Le jeu de données 3 semble avoir des scores de rappel plus faibles, ce qui peut suggérer qu'il est plus difficile pour les modèles de capturer le concept. Ceci est en accord avec le faible score d'acceptation dans la généralisation de nouveaux termes.

Par ailleurs, une analyse des termes non validés révèle la présence d'acronymes d'abréviations et de multi-termes imprécis ainsi que des termes à connotation plus "scientifiques" qui bien qu'utilisés dans des contextes de recherche, ne sont pas directement associés à l'innovation non-technologique.

## 7. Conclusion

Notre étude a démontré la capacité d'exploiter le Web comme une ressource pour la création d'un corpus multilingue riche et diversifié, indispensable à la compréhension de dynamique des connaissances. L'outil de *scrapping* que nous avons développé, adossé à *Scrapy*, démontre son efficacité dans le filtrage et l'extraction de données textuelles spécifiques.

Les performances de notre méthodologie, validées par des scores F1 élevés et des taux de rappel satisfaisants, confirment son utilité dans la reconnaissance de concepts émergents.

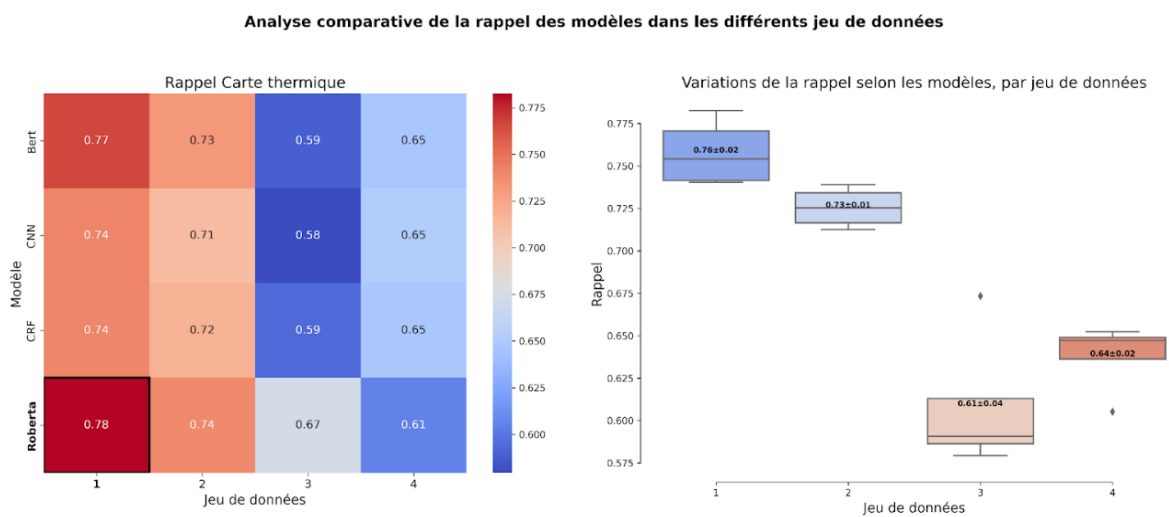

*Figure 2 : Rappel des modèles dans les différents jeux de données*

**Nombre de nouveaux termes générés (excluant les 1-grammes)**

| Modèle | Jdd 1 (%) | Jdd 2 (%) | Jdd 3 (%) | Jdd 4 (%) | % Moyenne |
|--------|-----------|-----------|-----------|-----------|-----------|
| CNN | 427 (20.14%) | 634 (18.45%) | 379 (21.37%) | 253 (26.09%) | 21.51% |
| CRF | 479 (21.29%) | 372 (23.66%) | 244 (28.69%) | 328 (25.00%) | 24.65% |
| BERT | 216 (44.90%) | 269 (44.98%) | 93 (41.95%) | 170 (51.76%) | 45.90% |
| RoBERTa | 202 (69.80%) | 234 (67.52%) | 182 (51.65%) | 503 (__77.73__%) | __66.68%__ |

*Tableau 4 : Nouveaux termes générés en excluant les 1-grammes. Chaque colonne correspond à un jeu de données (Jdd). Les pourcentages entre parenthèses sont les taux d'acceptation des nouveaux termes par les experts*

Néanmoins, le travail a rencontré des défis, notamment dans la manipulation des données multilingues et l'ajustement des seuils pour la sélection des termes, nous poussant à affiner nos techniques. En reconnaissant les limites de cette recherche, notamment la dépendance aux structures de données web et des variations de performances entre les modèles d'apprentissage, nous proposons des pistes pour des travaux futurs ; incluant l'intégration de sources de données supplémentaires l'évaluation de nouvelles architectures de modèles (comme LLM[10]), pour une extraction plus précise des termes, et une analyse sémantique contextuelle des termes. Cette étude ouvre la voie à une meilleure compréhension et analyse d'identification des termes liés à des concepts émergents.

## Bibliographie

---

[10] Large Language Model (Grand modèle de langage)